\documentclass[a4paper]{article}

\usepackage{INTERSPEECH2021}

\usepackage{cite}
\usepackage{tikz}
\usetikzlibrary{positioning}
\usepackage{tipa}
\usepackage{pgfplots}
\pgfplotsset{compat=1.9}
\usepackage{pifont}
\usepackage{enumitem}

\usepackage[T1]{fontenc}

\newcommand{\Hquad}{\hspace{0.0em}} 
\newcommand\mypar[1]{\noindent\textbf{#1}\Hquad}

\newcommand{\Sref}[1]{\S\ref{#1}}
\newcommand{\Fref}[1]{Figure~\ref{#1}}

\newcommand{\cmark}{\ding{51}}%
\newcommand{\xmark}{\ding{55}}%
\newcommand{\Tref}[1]{Table~\ref{#1}}

\title{Differentiable Allophone Graphs for Language-Universal Speech Recognition}
\name{Brian Yan, Siddharth Dalmia, David R. Mortensen, Florian Metze, Shinji Watanabe}
\address{Language Technologies Institute, Carnegie Mellon University, USA}
\email{\{byan,sdalmia\}@cs.cmu.edu}

\begin{document}
\ninept

\maketitle
\begin{abstract}
Building language-universal speech recognition systems entails producing phonological units of spoken sound that can be shared across languages.
While speech annotations at the language-specific phoneme or surface levels are readily available, annotations at a universal phone level are relatively rare and difficult to produce.
In this work, we present a general framework to derive phone-level supervision from only phonemic transcriptions and phone-to-phoneme mappings with \textit{learnable} weights represented using weighted finite-state transducers, which we call \textit{differentiable allophone graphs}.  
By training multilingually, we build a universal phone-based speech recognition model with interpretable probabilistic phone-to-phoneme mappings for each language.
These phone-based systems with learned allophone graphs can be used by linguists to document new languages, build phone-based lexicons that capture rich pronunciation variations, and re-evaluate the allophone mappings of seen language.
We demonstrate the aforementioned benefits of our proposed framework with a system trained on 7 diverse languages.
\end{abstract}
\noindent\textbf{Index Terms}: universal phone recognition, differentiable WFST, multilingual ASR, phonetic pronunciation, allophones

\section{Introduction}

The objective of language-universal speech recognition is to indiscriminately process utterances from anywhere in the world and produce intelligible transcriptions of what was said \cite{kohler2001multilingual, schultz2001language}.
In order to be truly universal, recognition systems need to encompass not only speech from many languages, but also intra-sentential code-switched speech \cite{bullock2009cambridge, li2019codeswitch}, speech with accents or otherwise non-standard pronunciations \cite{coupland2007style, sun2018domain}, and speech from languages without known written forms \cite{himmelmann2006language, hillis2019unsupervised}.

Language-universal speech recognition requires phonological units that are agnostic to any particular language such as articulatory features \cite{stuker2003integrating, livescu2016articulatory, li2020towards} or global phones \cite{schultz2002globalphone, li2020universal}, which can be annotated through examination of audio data.
While recent advancements in the related field of multilingual speech recognition have significantly improved the language coverage of a single system \cite{adams2019massively, pratap2020massively}, these works differ in that they operate on language-specific levels of surface vocabulary units \cite{li2019bytes} or phonemic units that are defined with reference to the unique phonological rules of each language \cite{dalmia2018sequence}. 
Prior works have avoided universal phone level annotation by implicitly incorporating this knowledge in shared latent representations that map to language-specific phonemes with neural nets \cite{stolcke2006cross, vesely2012language, dalmia2018sequence}. 

\begin{figure}[t]
\centering
    \begin{tikzpicture}[>=stealth, node distance=2mm and 8mm]%
    \node (8) {\textipa{[a]}};
    \node (9) [below=of 8] {\textipa{[a:]}};
    \node (10) [right=of 8] {\textipa{/a/}};
    \draw[->] (8) -- (10);
    \draw[->] (9) -- (10);
    \node[xshift=7mm] (11) [above=1mm of 8] {\textsc{Many-to-One}};

    \node (5) [right=of 10] {\textipa{[k]}};
    \node (6) [right=of 5] {\textipa{/k/}};
    \node (7) [below=of 6] {\textipa{/q/}};
    \draw[->] (5) -- (6);
    \draw[->] (5) -- (7);
    \node[xshift=7mm] (12) [above=1mm of 5] {\textsc{One-to-Many}};
    
    \node (1) [right=of 6] {\textipa{[s]}};
    \node (2) [right=of 1] {\textipa{/s/}};
    \node (3) [below=of 1] {\textipa{[S]}};
    \node (4) [right=of 3] {\textipa{/S/}};
    \draw[->] (1) -- (2);
    \draw[->] (1) -- (4);
    \draw[->] (3) -- (2);
    \draw[->] (3) -- (4);
    \node[xshift=7mm] (13) [above=1mm of 1] {\textsc{Many-to-Many}};
\end{tikzpicture}%
    \vspace{-15pt}
    \caption{Examples showing three types of manifold mappings of phones (in brackets) to phonemes (in slashes). Many-to-one describes allophones of a phoneme. One-to-many describes a duplicitous phone that maps to multiple phonemes. Many-to-many consists of both allophones and duplicitous phones.}
    \label{fig:manifold}
    \vspace{-14pt}
\end{figure}
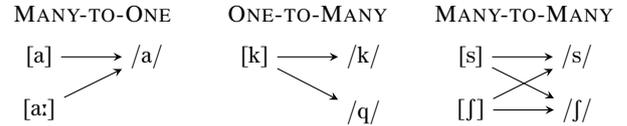

Another approach is to learn explicit universal phone representations by relating language-specific units to their universal phonetic distinctions.
Instead of relying on phone annotations, these prior works approximate universal phonological units through statistical acoustic-phonetic methods \cite{kohler2001multilingual} or phone-to-phoneme realization rules \cite{mortensen-etal-2020-allovera, li2020universal}.
Unlike the implicit latent approach, this method allows for language-universal prediction. However, performance is dependent on the clarity of phone-phoneme dynamics in the selected training languages \cite{kohler1996multi, li2020universal}.

We are interested in systems that can incorporate the strengths of both the implicit and explicit approaches to representing universal phones.
In particular, we are interested in language-universal automatic speech recognition (ASR) systems that can
1) explicitly represent universal phones and language-specific phonemes,
2) be built using only automatically generated grapheme-to-phoneme annotations and phone-to-phoneme rules,
3) resolve naturally ambiguous phone-to-phoneme mappings using information from other languages,
and 4) learn interpretable probabilistic weights of each mapping.

In this work, we seek to incorporate these desiderata in a phone-based speech recognition system. 
We first propose a general framework to represent phone-to-phoneme rules as \textit{differentiable allophone graphs} using weighted finite-state transducers \cite{mohri2002weighted, moritz2020semi, doetsch2017returnn, shao2020pychain, hannun2020differentiable, k2} to probabilistically map phone realizations to their underlying language-specific phonemes (\Sref{sec:encodingwfst}). 
We then incorporate these differentiable allophone graphs in a multilingual model with a universal phone recognizing layer trained in an end-to-end manner, which we call the AlloGraph model (\Sref{sec:allograph}).
We show the efficacy of the AlloGraph model in predicting phonemes for 7 seen languages and predicting phones for 2 unseen languages with comparison to prior works (\Sref{sec:results}). 
More importantly, we show that our model resolves the ambiguity of manifold phone-to-phoneme mappings with an analysis of substitution errors and an examination of the interpretable allophone graph weights (\Sref{sec:disambiguate}).
Finally we demonstrate our phone-based approach in two linguistic applications: pronunciation variation and allophone discovery (\Sref{sec:applications}).

\section{Background and Motivation}

In this section, we first introduce phone-to-phoneme mappings for manufacturing phone supervision from phoneme annotations (\Sref{units}). Then we discuss short-comings of a baseline method representing mappings as a pass-through matrix (\Sref{sec:allomat}) to motivate our graph-based framework in the subsequent section (\Sref{sec:proposed}). 

\subsection{Phonological Units}
\label{units}

\subsubsection{Language-Specific Phonemes vs. Universal Phones}
A phone $n$ is a unit of spoken sound within a universal set $\mathcal{N}$ which is invariant across all languages, where $\mathcal{N} = \{ n_1, ..., n_{|N|} \}$ consists of $|\mathcal{N}|$ total phones \cite{schultz2002globalphone}. In contrast, a phoneme $m^{(l)}$ is a unit of linguistically contrastive sound for a given language $l$ within a language specific set, where $\mathcal{M}^{(l)} = \{ m_1^{(l)}, ..., m_{|\mathcal{M}^{(l)}|}^{(l)} \}$ consists of $|\mathcal{M}^{(l)}|$ total phonemes \cite{mortensen2018epitran}. 
Phonemes defined for different languages describe different underlying sounds. Multilingual systems that conflate phonemes across languages have been shown to perform worse than those that treat phonemes as language-specific \cite{kohler1996multi, li2020universal}. 

\subsubsection{Phone-to-Phoneme Mappings}
\label{sec:mappings}

For each language, the phone-to-phoneme mappings are defined as a series of tuples, $(n_i, m_j^{(l)})$, where $m_j^{(l)} \in \mathcal{M}^{(l)}$ and $n_i \in \mathcal{N'} \subseteq \mathcal{N}$ for some subset $\mathcal{N'}$ of phones that occur as realizations in the language. Each phoneme has one or more phone realization and not all universal phones are necessarily mapped to a phoneme grounding in a particular language. Note that mappings may be imperfect in our resources \cite{mortensen-etal-2020-allovera}.

Phone-to-phonemes can be one-to-one mappings, but often the relationships are manifold.
As shown in \Fref{fig:manifold}, many-to-one mappings are found in scenarios where multiple phones are allophones, or different realizations, of the same phoneme. This is the prototypical mapping type. One-to-many mappings also occur for duplicitous phones that are mapped to multiple phonemes.\footnote{These occur in resources like \cite{mortensen-etal-2020-allovera} when the source conflates allophonic and morphophonemic alternations,
in instances of archiphonemic underspecification and neutralization (e.g. treating Japanese [m] as a realization of both /m/ and /N/ or English [\textfishhookr] as a realization of both /t/ and /d/ as in \textit{writer} \textipa{[\textturnr aj\textfishhookr\textturnr]} and \textit{rider} \textipa{[\textturnr a:j\textfishhookr\textturnr]}), or---spuriously---when the grapheme-phoneme mapping is complex.}
Furthermore, many-to-one and one-to-many mappings can occur together in various many-to-many forms. 

\subsubsection{Manufacturing Phone-Level Supervision}

Since phones are fine-grained distinctions of spoken sounds in the universal space, phonemes are only fuzzy approximations. Multilingual sharing between diverse languages is required to properly learn phonetic distinctions. Consider the following:

\mypar{One-to-One:} If a phone is mapped one-to-one with a phoneme, then the learned phone representation will directly correspond to one supervising phoneme. In the multilingual setting, these direct mappings help other languages disambiguate this phone.

\mypar{One-to-Many:} If a phone is mapped to many phonemes, then each phoneme provides supervision in proportion to their prior distributions. 
If the learned phonemes representations are mapped from the learned phone, phoneme confusions occur if the one-to-many mappings are not disambiguated.
This ambiguity persists despite information sharing from other languages.

\mypar{Many-to-One:} If many phones are mapped to a phoneme, each phone receives the same supervision. A second language with complementary mappings is required to learn distinct phones. 

\mypar{Many-to-Many:} When one-to-many and many-to-one mappings occur together, they can take various forms. Generally, the many-to-one portions can be resolved through multilingual sharing but the one-to-many portions would still be problematic.

\subsection{Encoding Phone-to-Phoneme as Pass-through Matrix}
\label{sec:allomat}
Prior works have shown that phone-to-phoneme mappings can be encoded as pass-through layers that convert a phone distribution into a phoneme distribution \cite{li2020universal}. This phone-to-phoneme encoding, which we call AlloMatrix, is a sparse matrix $A^{(l)} = \{0, 1\}^{|\mathcal{N}| \times |\mathcal{M}^{(l)}|}$ where each $(n_i, m_j^{(l)})$ tuple in the mappings desribed in \Sref{sec:mappings} is represented by $a^{(l)}_{i,j} = 1$. The AlloMatrix transforms a logit vector of phones, $\mathbf{p}^{\mathcal{N}} = [p^{\mathcal{N}}_i, ..., p^{\mathcal{N}}_{|\mathcal{N}|}]$, to a logit vector of phonemes, $\mathbf{p}^{\mathcal{M}^{(l)}} = [p^{\mathcal{M}^{(l)}}_j, ..., p^{\mathcal{M}^{(l)}}_{|\mathcal{M}^{(l)}|}]$ by the dot product of the $j$th column of $A^{(l)}$ with each phone logit $p_i^\mathcal{N}$:
  \vspace{-4pt}
\begin{equation}
p^{\mathcal{M}^{(l)}}_j = \sum_{i}^{|\mathcal{N}|}{(a^{(l)}_{i,j}) (p_i^\mathcal{N}) } \label{eq:allomat}
  \vspace{-1pt}
\end{equation}
In the many-to-one approach, this amounts to summing the phone contributions which is in accordance with our desired mapping of allophones in \Sref{sec:mappings}.
However, in one-to-many mappings a phone logit broadcast equally to each of the phonemes. This disagrees with the definition of phone realization. Rather we state that a realized phone in an utterance is grounded to each of the mapped phonemes with probability.

\section{Proposed Framework}
\label{sec:proposed}
\subsection{Encoding Phone-to-Phoneme as WFST}
\label{sec:encodingwfst}

We define the allophone graph for language $l$, denoted by $G^{(l)}$, to be a single state weighted finite-state transducer (WFST) with a transition function $\pi(n_i,  m_j^{(l)})$ giving each phone-to-phoneme mapping and a corresponding weight function $w(n_i,  m_j^{(l)})$ giving the likelihood that $n_i$ is the phonetic realization of $m_j^{(l)}$ for each transition. 
The allophone graph $G^{(l)}$ accepts phone emission probabilities $E^\mathcal{N}$ and transduces them into phonemes $E^{\mathcal{M}^{(l)}}$ through WFST composition \cite{mohri2002weighted}, which is denoted as $\circ$.
\begin{equation}
E^{\mathcal{M}^{(l)}} = E^\mathcal{N} \circ G^{(l)} \label{eq:allograph}
\end{equation}

This WFST is an analogous data structure to the aforementioned matrix in \Sref{sec:allomat}, but this graphical representation of phone-to-phoneme mappings as arcs in a probabilistic transduction allows us to make two key intuitive determinations.
First, many-to-one mappings are transductions of several phones into the same phoneme and therefore the phoneme posterior is given by summing over the input phone posteriors, as is also done in \Sref{sec:allomat}.
Second, one-to-many mappings are transductions splitting the posterior of a single phone to several phoneme posteriors, depending on how likely those phonemes are to be groundings of the phone. In \Sref{sec:allomat}, the broadcasting method fails to do this probabilistic splitting in one-to-many scenarios, creating ambiguity.

\subsection{Phone Recognition with Allophone Graphs}
\label{sec:allograph}

In this section, we apply the allophone graphs as differentiable WFST \cite{mohri2002weighted, moritz2020semi, doetsch2017returnn, shao2020pychain, hannun2020differentiable, k2} layers in phone-based ASR systems optimized with only multilingual phoneme supervision.

\begin{table*}[t]
  \centering
    \caption{Results presenting the performances of our proposed AlloGraph models with our implementations of Phoneme-Only and AlloMatrix baselines, as measured by language-specific phoneme error-rate (\%) for seen languages and universal phone error-rate (\%) for unseen languages. Performances on unseen languages were evaluated using phone-level annotations for the Tusom and Inuktitut corpora. Note that while our proposed AlloGraph and our baseline AlloMatrix models produce both phone and phoneme-level predictions, the Phoneme-Only approach only recognizes language-specific phonemes. The averaged totals across unseen/seen are shown in \textbf{bold} and the best performing models in each category are shown in \underline{\textbf{bold}}.
    }
    \label{tab:main_results}
  \vspace{-5pt}
    \resizebox {\linewidth} {!} {
\begin{tabular}{llccccccccc|ccc}
\toprule
& & Uses & \multicolumn{8}{c}{Seen (Phoneme Error Rate \%)} & \multicolumn{3}{c}{Unseen (Phone Error Rate \%)} \\
\cmidrule(r){4-11}\cmidrule(r){12-14}
Model Type & Model Name & Phones & Eng & Tur & Tgl & Vie & Kaz & Amh & Jav & Total & Tusom & Inuktitut & Total \\
\midrule
Phoneme-Only & Multilingual-CTC~\cite{dalmia2018sequence} & \xmark & 25.3 & 27.7 & 28.5 & 31.9 & 31.5 & 28.6 & 35.2 & \underline{\textbf{29.8}} & \multicolumn{3}{c}{\textit{No Phone Predictions}} \\
\midrule
AlloMatrix & Allosaurus \cite{li2020universal} & \cmark & 26.5 & 27.6 & 33.1 & 32.0 & 31.9 & 28.2 & 39.0 & \textbf{31.2} & 91.2 & 96.7 &\textbf{94.0} \\ 
AlloGraph & Our Proposed Model & \cmark & 26.0 & 28.6 & 28.2 & 31.9 & 32.5 & 29.1 & 36.2 & \underline{\textbf{30.5}} & 81.2 & 85.8 & \textbf{84.1} \\
AlloGraph & + Universal Constraint (UC) & \cmark & 27.3 & 28.7 & 29.9 & 32.5 & 35.1 & 30.9 & 36.6 & \textbf{31.6} & 80.5 & 79.9 & \underline{\textbf{80.2}} \\ 
\bottomrule
\end{tabular}
}

  \vspace{-10pt}
\end{table*}

In this work, we use the connectionist temporal classification network (CTC) \cite{graves2006connectionist, miao2015eesen} where a language-universal $\textsc{Encoder}$ maps input sequence $\mathbf{x} = [\mathbf{x}_t, ..., \mathbf{x}_T]$ to a sequence of hidden representations $\mathbf{h} = [\mathbf{h}_t, ..., \mathbf{h}_T]$, where $\mathbf{h}_t \in \mathbb{R}^{d}$. 
The phone emission probabilities $E^\mathcal{N\cup \varnothing}$ are given by the affine projection of $\mathbf{h}$ followed by the softmax function, denoted as \textsc{SoftmaxOut}.\footnote{In training, logits corresponding to unmapped phones in a particular language are masked prior to being softmax normalized similar to \cite{dalmia2019plm}.}
To handle the blank token $\varnothing$ used in CTC to represent the null emission \cite{graves2006connectionist}, we add the $\varnothing \rightarrow \varnothing$ transition as an additional arc in the language-specific allophone graphs $G^{(l)}$. 
Phone and phoneme emissions are thus given by:
  \vspace{-2pt}
\begin{align}
  \vspace{-6pt}
\mathbf{h} &= \textsc{Encoder}(\mathbf{x}) \label{eq:enc} \\
E^{\mathcal{N} \cup \varnothing} &= \textsc{SoftmaxOut}(\mathbf{h}) \label{eq:out} \\
E^{\mathcal{M}^{(l)} \cup \varnothing} &= E^{\mathcal{N} \cup \varnothing} \circ G^{(l)} \label{eq:diffwfst}
\end{align}
Equation \ref{eq:diffwfst} shows the CTC specific form of the general phone-to-phoneme emission transduction shown in Equation \ref{eq:allograph}.
During training, we maximize the likelihood of the ground-truth phonemes $y = [{y}_1, ..., {y}_S]$, where ${y}_s \in \mathcal{M}^{(l)}$ and $S$ is the length of the ground-truth which is at most the length of the input $T$, by marginalizing over all possible CTC alignments using the forward-backward computation \cite{graves2006connectionist, miao2015eesen}.

We refer to this multilingual CTC architecture with allophone graphs as \textit{our proposed AlloGraph} model. In the vanilla AlloGraph, we allow the weights of $G^{(l)}$ to freely take on any values. This is a loose-coupling of phone and phoneme emissions where each $G^{(l)}$ may amplify or reduce the phone posteriors; for instance, this allows $G^{(l)}$ to learn cases where a phone is universally rare but is a prominent realization in language $l$. 

While loose-coupling of phone and phoneme emissions is beneficial to language-specific phoneme recognition, it dilutes supervision to the universal phone layer. We address this by enforcing a tight-coupling of phone and phoneme emissions such that the phone posterior is only isometrically transformed: $\sum_{m^{(l)} \in \mathcal{M}'^{(l)}} w(n_i, m) = 1$, where $\mathcal{M}'^{(l)}$ is the subset of phonemes $\mathcal{M}^{(l)}$ that $n_i$ is mapped to in language $l$. 
Now, Equation \eqref{eq:diffwfst} exactly sums phone posteriors for many-to-one and splits phone posteriors for one-to-many in the manner that we desire, as stated in \Sref{sec:encodingwfst}. 
We call this tightly-coupled variant the \textit{AlloGraph + Universal Constraint (UC)} model.

\begin{table}[t]
  \centering
    \caption{Results showing the performance of the AlloMatrix and AlloGraph models on two unseen language, as measured by Phone Error Rate (PER), Substitution Error Rate (SER), and Articulatory Feature Distance (AFD). AFD measures the severity of substitution errors, computed via the distance between vectors of 22 articulatory features corresponding to each phone.}
    \label{tab:panphon}
  \vspace{-5pt}
    \resizebox {\linewidth} {!} {
\begin{tabular}{lcccccc}
\toprule
& \multicolumn{3}{c}{Tusom} & \multicolumn{3}{c}{Inuktitut} \\
\cmidrule(r){2-4} \cmidrule(r){5-7}
Model & PER & SER & AFD & PER & SER & AFD \\
\midrule
AlloMatrix & 91.2 & 65.6 & 12.3 & 96.7 & 75.3 & 12.4 \\
\midrule
AlloGraph & 81.2 & 56.8 & 8.7 & 85.8 & 65.8 & 8.4 \\
+ UC & \textbf{80.5} & \textbf{54.9} & \textbf{7.8} & \textbf{79.9} & \textbf{59.9} & \textbf{7.8} \\
\bottomrule
\end{tabular}
}
  \vspace{-10pt}
  \label{tab:unseen_results}
\end{table}

\section{Data and Experimental Setup}
\label{sec:setup}
\mypar{Data:} 
We use the English LDC Switchboard Dataset \cite{godfrey1993switchboard, ldceval2kctm, ldceval2kspeech} and 6 languages from the IARPA BABEL Program: Turkish, Tagalog, Vietnamese, Kazakh, Amharic and Javanese~\cite{babel}.
These datasets contain 8kHz recordings of conversational speech each containing around 50 to 80 hours of training data, with an exception of around 300 hours for English. We also consider two indigenous languages with phone level annotations, Tusom \cite{mortensen2021tusom2021} and Inukitut, during evaluation only.
We obtain phonemic annotations using Epitran for auto grapheme-to-phoneme \cite{mortensen2018epitran} and phone-to-phoneme rules from Allovera \cite{mortensen-etal-2020-allovera}. 

\mypar{Experimental Setup:} All our models were trained using the ESPnet toolkit \cite{watanabe2018espnet} with differentiable WFSTs implemented using the GTN toolkit \cite{hannun2020differentiable}.
To prepare our speech input features we first upsample the audio to 16kHz, augment it by applying a speed perturbation of $0.9$ and $1.1$, and then extract global mean-variance normalized 83 log-mel filterbank and pitch features. Input frames are processed by an audio encoder with convolutional blocks to subsample by $4$ \cite{watanabe2018espnet} before feeding to $12$ transformer-encoder blocks with a feed-forward dim of $2048$, attention dim of $256$, and $4$ attention heads. 
We augment our data with the Switchboard Strong (SS) augmentation policy of SpecAugment \cite{specaug} and apply a dropout of $0.1$ for the entire network. 
We use the Adam optimizer to train 100 epochs with an inverse square root decay schedule, a transformer-lr scale~\cite{watanabe2018espnet} of $5$, $25$k warmup steps, and an effective batchsize of $768$.

\section{Results}
\label{sec:results}

In \Tref{tab:main_results}, we show the results of our AlloGraph and AlloGraph + UC models. 
As mentioned in \Sref{sec:setup}, we use Tusom and Inuktitut as two unseen languages with phone level annotations to evaluate our language-universal predictions; since these languages are unseen our model does not know their phoneme sets or which phones appear as realizations, allowing us to assess how universal our phone-based predictions are.
On these two unseen languages our AlloGraph model outperforms our AlloMatrix baseline based on \cite{li2020universal} by an average of 9.9 phone error-rate (\%).
When using the Universal Constraint described in \Sref{sec:allograph}, our approach gains an additional 3.9 phone error-rate improvement. 
The AlloGraph models make fewer substitution errors than the AlloMatrix baseline, and the substitutions are also less severe; we examine these improvements further in \Sref{sec:afd}.

\Tref{tab:main_results} also shows the language-specific phoneme level performance of the AlloGraph model on 7 seen languages. 
Note that these languages are annotated with phonemes as described in \Sref{sec:setup} but not with phones.
Here our AlloGraph model slightly outperforms the AlloMatrix baseline, but both show degradation compared to our Phoneme-Only\footnote{Phoneme-Only \cite{dalmia2018sequence} directly maps the shared \textsc{Encoder} hidden states to language-specific phoneme level \textsc{SoftmaxOut}, replacing the shared phone level in Equation \eqref{eq:out}. Thus there are no phone predictions.} baseline based on \cite{dalmia2018sequence}. 
We observe that models placing emphasis on learning universal phones do so with some cost to the language-specific level. 

The AlloGraph is advantageous in jointly modeling phones and phonemes compared to the AlloMatrix baseline due to learned disambiguations of phone-to-phoneme mappings; we examine this benefit further in \Sref{sec:disambiguate}.  

\subsection{Universal Phone Recognition for Unseen Languages}
\label{sec:afd}

\begin{table}[t]
  \centering
    \caption{Results showing the top 3 phone confusion pairs of the AlloMatrix and AlloGraph + UC models on two unseen languages. Confusion pairs are denoted as [correct] $\rightarrow$ [incorrect]. Articulatory Feature Distance (AFD) measures the severity of each confusion, computed via the distance between vectors of 22 articulatory features corresponding to each phone.}
    \label{tab:confusions}
  \vspace{-5pt}
    \resizebox {\linewidth} {!} {
\begin{tabular}{lcccccc}
\toprule
& \multicolumn{2}{c}{Tusom} & \multicolumn{2}{c}{Inuktitut} \\
\cmidrule(r){2-3} \cmidrule(r){4-5}
Model & Confusion & AFD & Confusion & AFD \\
\midrule
 & \textipa{[\textbari]} $\rightarrow$ \textipa{[\textlowering{\textbeta}]} & 15 & \textipa{[a]} $\rightarrow$ \textipa{[\textlowering{\textbeta}]} & 13 \\
AlloMatrix & \textipa{[\textschwa]} $\rightarrow$ \textipa{[\textlowering{\textbeta}]} & 13 & \textipa{[i]} $\rightarrow$ \textipa{[\textlowering{\textbeta}]} & 13 \\
 & \textipa{[\textschwa]} $\rightarrow$ \textipa{[s']} & 17 & \textipa{[u]} $\rightarrow$ \textipa{[s']} & 23 \\
\midrule
 & \textipa{[i]} $\rightarrow$ \textipa{[i:]} & 2 & \textipa{[a]} $\rightarrow$ \textipa{[\=*A]} & 3 \\
AlloGraph & \textipa{[k]} $\rightarrow$ \textipa{[\texttoptiebar{kp}]} & 4 & \textipa{[u]} $\rightarrow$ \textipa{[o]} & 4 \\
 & \textipa{[a]} $\rightarrow$ \textipa{[a:]} & 2  & \textipa{[a]} $\rightarrow$ \textipa{[a:]} & 2 \\
 \midrule
 & \textipa{[a]} $\rightarrow$ \textipa{[\textturna]} & 4 & \textipa{[q]} $\rightarrow$ \textipa{[k]} & 2 \\
AlloGraph + UC & \textipa{[\textschwa]} $\rightarrow$ \textipa{[\textturna]} & 2 & \textipa{[a]} $\rightarrow$ \textipa{[\textturna]} & 4 \\
 & \textipa{[a]} $\rightarrow$ \textipa{[A]} & 2  & \textipa{[i]} $\rightarrow$ \textipa{[I]} & 2 \\
\bottomrule
\end{tabular}
}
  \vspace{-10pt}
\end{table}

As shown in \Tref{tab:unseen_results}, the improvements of the AlloGraph models over the AlloMatrix baseline come from reduced phone substitution errors.
In addition to making fewer substitution errors, the AlloGraph models also make less severe substitutions than the AlloMatrix baseline.
We quantify this severity by computing the averaged distance between articulatory feature vectors \cite{mortensen-etal-2016-panphon} between the ground truth and incorrectly predicted phones for all substitution errors.
Compared to the AlloMatrix, the substitutions made by the AlloGraph and AlloGraph + UC models are 31\% and 37\% closer in articulatory feature distance (AFD).

The high AFD of the AlloMatrix baseline results from degenerate behavior in which vowels are frequently confused for plosives, as shown by the top confusion pairs in \Tref{tab:confusions}.
On the other, the top confusion pairs of the AlloGraph models are between related vowels which are proximate in the articulatory feature space.
Thus the AlloGraph models produce intelligible phone transcriptions, while the AlloMatrix model fails.  
For qualitative examples of phone recognition, please see \Sref{sec:qualitative}.

\subsection{Probabilistic Phone-to-Phoneme Disambiguation}
\label{sec:disambiguate}
\label{sec:interpretable}

\begin{figure}[t]
\centering
    \begin{tikzpicture}[>=stealth, node distance=6mm and 14mm, auto]%
    \node (5) [right=of 10] {\textipa{[k]}};
    \node (6) [right=of 5] {\textipa{/q/}};
    \node (7) [below=of 6] {\textipa{/k/}};
    \draw[->, line width=0mm, dashed] (5) -- node[anchor=south] {$0.0$} (6);
    \draw[->, line width=0.8mm] (5) -- node[anchor=north, xshift=-2ex] {$1.0$} (7);
    \node[xshift=10mm] (12) [above=1mm of 5] {\textsc{One-to-Many}};
    
    \node (1) [right=of 6] {\textipa{[s]}};
    \node (2) [right=of 1] {\textipa{/s/}};
    \node (3) [below=of 1] {\textipa{[S]}};
    \node (4) [right=of 3] {\textipa{/S/}};
    \draw[->, line width=0.8mm] (1) -- node[anchor=south] {$1.0$} (2);
    \draw[->, line width=0mm, dashed] (1) -- node[rotate=-30, xshift=1ex] {$0.0$} (4);
    \draw[->, line width=0.6mm] (3) -- node[rotate=30, xshift=-1ex] {$0.75$} (2);
    \draw[->, line width=0.2mm] (3) -- node[anchor=north] {$0.25$} (4);
    \node[xshift=10mm] (13) [above=1mm of 1] {\textsc{Many-to-Many}};
    \node[xshift=10mm] (14) [below=1mm of 3] {\textsc{(Tagalog)}};
    \node[xshift=-32mm] (15) [below=1mm of 3] {\textsc{(Javanese)}};
\end{tikzpicture}%
        \vspace{-15pt}
    \caption{Examples of disambiguated phone-to-phoneme mappings using the interpretable weights of our AlloGraph + UC model, where each [phone] is probabilistically mapped to a /phoneme/. In the one-to-many example from Javanese, [k] is predominantly a realization of /k/. In the many-to-many example from Tagalog, [s] is predominantly a realization of /s/ while \textipa{[S]} is a realization of /s/ 75\% of the time and \textipa{/S/} otherwise.}
    \label{fig:priors}
    \vspace{-10pt}
\end{figure}
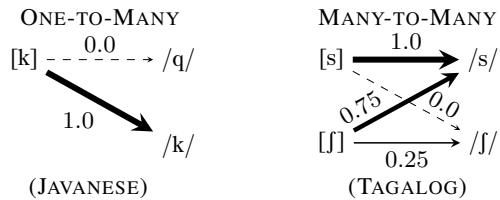

An added benefit of our model is the ability to interpret the weights of learned AlloGraphs, which show disambiguations of ambiguous phone-to-phoneme mappings. As shown in \Fref{fig:priors}, our AlloGraph + UC model distributes phone emissions to multiple phonemes in the one-to-many and many-to-many scenarios. These probabilities can be interpreted as prior distributions of each mapping captured by the allophone graph and can be used to determine the relative dominance of each arc in manifold mappings that can be otherwise difficult to explain.

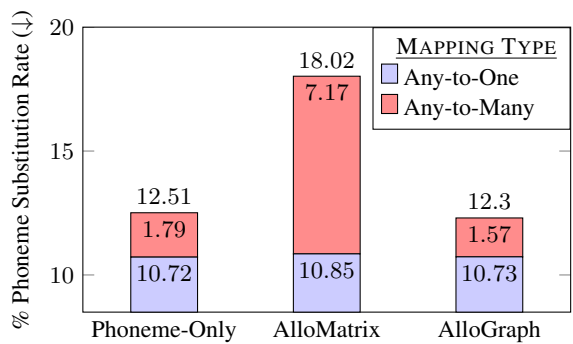
\begin{figure}[t]
\centering
  \begin{tikzpicture}
  
      \pgfplotsset{
        show sum on top/.style={
            /pgfplots/scatter/@post marker code/.append code={%
                \node[
                    at={(normalized axis cs:%
                            \pgfkeysvalueof{/data point/x},%
                            \pgfkeysvalueof{/data point/y})%
                    },
                    anchor=south,
                    text=black,
                ]
                {\pgfmathprintnumber{\pgfkeysvalueof{/data point/y}}};
            },
        },
    }
    
\begin{axis}[
    ybar stacked, ymin=0,  
    bar width=25pt,
    width=\linewidth,
    height=0.67\linewidth,
    ymax=20,
    ymin=8.5,
    enlarge x limits=0.25,
    symbolic x coords={Phoneme-Only,AlloMatrix,AlloGraph},
    xtick=data,
    nodes near coords, 
    nodes near coords align={anchor=north},
    nodes near coords bar offset=1,
    every node near coord/.style={},
    legend cell align={left},
    ylabel={\% Phoneme Substitution Rate ($\downarrow$)},
    legend style={at={(1,1)},anchor=north east},
    xtick style={draw=none},
  ]
  \addlegendimage{empty legend}\addlegendentry{\hspace{-.1cm}\underline{\textsc{Mapping Type}}}
  \addplot [fill=blue!20,fill opacity=1.0,text=black] coordinates {
({Phoneme-Only},10.72)
({AlloMatrix},10.85)
({AlloGraph},10.73)}; \addlegendentry{Any-to-One}
  \addplot [fill=red!45,fill opacity=1.0,text=black, show sum on top] coordinates {
({Phoneme-Only},1.79)
({AlloMatrix},7.17)
({AlloGraph},1.57)}; \addlegendentry{Any-to-Many}
  \end{axis}
  \end{tikzpicture}
    \vspace{-15pt}
    \caption{Results comparing the performances of our baseline Phoneme-Only, baseline AlloMatrix, and proposed AlloGraph models on a high phone-to-phoneme complexity language, Tagalog, as measured by phoneme substitution error-rate (\%). The any-to-one category includes phonemes in one-to-one and many-to-one mappings, and any-to-many includes phonemes in one-to-many and many-to-many mappings.}
    \label{fig:confusions}
        \vspace{-15pt}
\end{figure}

The performance of AlloGraph + UC on languages with complex phone-phoneme mappings, such as Tagalog and Javanese, is greatly improved over the AlloMatrix baseline. 
In these languages, phones are frequently defined as realizations of multiple ostensive phonemes and there are many allophones of each phoneme. 
As shown in \Fref{fig:confusions} these ambiguous mappings are especially detrimental to the AlloMatrix model, 
which produces a high number of phoneme substitution errors compared to our AlloGraph model and Phoneme-Only baseline.

\subsection{Linguistic Applications}
\label{sec:applications}
In this section, we demonstrate the efficacy of phone-based predictions from our AlloGraph + UC model in two applications.

As shown in \Tref{fig:pronunciations}, our AlloGraph + UC model produces different phonetic realizations of a single phonemic pronunciation. 
By collecting all of the phonetic realizations for correct phonemic transcriptions of the word `hello' uttered by numerous speakers across test sets in our conversational corpora, we automatically identified the most frequent phonetic pronunciations.
These qualitative examples suggest that dynamic methods for building lexicons using universal phone recognition systems can capture diverse pronunciations that can bolster knowledge sets \cite{coupland2007style}.
This may benefit pronunciation-sensitive tasks like code-switched \cite{li2019codeswitch} or accented speech recognition \cite{viglino2019end}.

\begin{table}[t]
\centering
    \caption{Results showing the pronunciations of the word `hello' across the 7 languages discovered by our AlloGraph + UC model, as shown in phonemic and phonetic forms. Pronunciation variations between different speakers in our conversational test set are captured at the phonetic level. We present the 3 most frequent phone-based pronunciations and their percentages.}
    \label{fig:pronunciations}
      \vspace{-5pt}
    \resizebox {\linewidth} {!} {
\begin{tabular}{lll|lr|lr|lr}
\toprule
& & \multicolumn{7}{c}{Pronunciations} \\
\cmidrule(r){3-9}
Lang. & Word & \multicolumn{1}{c}{Phonemic} & \multicolumn{6}{c}{Phonetic} \\
\midrule
Eng & hello & \textipa{/h\textschwa low/} & \textipa{[halo]} & 54\% & \textipa{[h@low]} & 8\% & \textipa{[hElow]} & 8\% \\
Tur & alo & \textipa{/alo/} & \textipa{[a:\textltilde o]} & 100\% & \multicolumn{2}{c}{-} & \multicolumn{2}{|c}{-} \\
Tgl & hello & \textipa{/hello/} & \textipa{[hello]} & 99\% & \textipa{[hellu]} & 1\% & \multicolumn{2}{c}{-} \\
Vie & a l\^{o} & \textipa{/\textglotstop a lo/} & \textipa{[\textglotstop a lo]} & 100\% & \multicolumn{2}{c}{-} & \multicolumn{2}{|c}{-} \\
Kaz & \includegraphics[height=0.22cm]{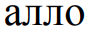} & \textipa{/Allo/} & \textipa{[\=*A\textsubbridge{l}\textsubbridge{l}o]} & 75\% & \textipa{[\=*A\textturnscripta \textsubbridge{l}\textsubbridge{l} o]} & 20\% & \textipa{[\textturnscripta\textsubbridge{l}\textsubbridge{l} o]} & 5\% \\
Amh & \includegraphics[height=0.22cm]{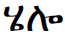} & \textipa{/helo/} & \textipa{[\texthth elo]} & 99\% & \textipa{[helo]} & 1\% & \multicolumn{2}{c}{-}\\
Jav & halo & \textipa{/halo/} & \textipa{[halo]} & 88\% & \textipa{[h\textopeno lo]} & 11\% & \textipa{[helo]} & 1\% \\
\bottomrule
\end{tabular}
}
\end{table}

\begin{table}[t]
\centering
    \caption{Results showing the most frequent triphone contexts and realization rates of various phones mapped to the phonemes /b/ and \textipa{/\textschwa/} in Amharic, as discovered by our AlloGraph + UC model on our test corpus. Phones that are not mapped to any phoneme, such as \textipa{[\textturna]} in Amharic, can still appear as hypothesized realizations suggesting new phone-to-phoneme mappings.}
    \label{fig:context}
    \resizebox {\linewidth} {!} {
\begin{tabular}{lcclll}
\toprule
Phone-to- & Realization & Predefined & \multicolumn{3}{c}{Frequent} \\
Phoneme & Rate (\%) & Mapping & \multicolumn{3}{c}{Triphone Contexts} \\
\midrule
\textipa{[b]} $\rightarrow$ \textipa{/b/} & 64.5 & \cmark & [\#\textipa{b\textturna]} & [\#\textipa{b\textschwa]} &  [\#\textipa{bI]} \\

\textipa{[\textlowering{\textbeta}]} $\rightarrow$ \textipa{/b/} & 29.7 & \cmark & \textipa{[\textopeno \textlowering{\textbeta}e]} & \textipa{[\textschwa\textlowering{\textbeta}\texthth]} &  [\#\textipa{\textlowering{\textbeta}I]} \\

\midrule

\textipa{[\textschwa]} $\rightarrow$ \textipa{/\textschwa/} & 32.7 & \cmark & \textipa{[n\textschwa w]} & \textipa{[d\textschwa \texthth]} &  \textipa{[d\textschwa t]} \\

\textipa{[\textturna]} $\rightarrow$ \textipa{/\textschwa/} & 29.2 & \xmark & \textipa{[\textglotstop \textturna l]} & \textipa{[s\textturna l]} &  \textipa{[s\textturna m]} \\

\textipa{[E]} $\rightarrow$ \textipa{/\textschwa/} & 16.4 & \cmark & \textipa{[gEr]} & \textipa{[bEr]} &  \textipa{[lEt]} \\

\textipa{[\textopeno]} $\rightarrow$ \textipa{/\textschwa/} & 13.8 & \cmark & \textipa{[\textglotstop \textopeno w]} & \textipa{[\textglotstop \textopeno j]} &  \textipa{[\textglotstop \textopeno n]} \\

\bottomrule
\end{tabular}
}
      \vspace{-5pt}
\end{table}

Since the AlloGraph + UC model produces joint alignments of phones and phonemes for seen languages, it can also discover the allophone realization rates and triphone contexts in test corpora (\Tref{fig:context}). 
Our method can also hypothesize new allophones such as the the phone \textipa{[\textturna]} which is not mapped to any of the phonemes in Amharic \cite{mortensen-etal-2020-allovera}. 
One important step in language documentation is discovering and defining the relationship between phones and phonemes \cite{himmelmann2006language}, ensuring that mappings are exhaustive but devoid of spurious pairs.
Automatic, data-driven methods to generate phone-phoneme mappings allow linguists to discover these relationships more effectively.

\section{Conclusion and Future Work}

We present differentiable allophone graphs for building universal phone-based ASR using only language-specific phonemic annotations and phone-to-phoneme rules. 
We show improvements in phone and phoneme prediction over prior works.
More importantly, our framework enables model interpretability and unique linguistic applications, such as phone-based lexicons and allophone discovery. 
In future work, we will seek to incorporate contexually dynamic phone-to-phoneme mappings using convolutional or attention-based WFST weights.
We hope that the insights of this work stimulate research on learnable representations of other linguistic rules, such as articulatory features \cite{li2020towards}, phonotactics \cite{feng2020phonotactics}, and cross-lingual mappings \cite{Hu2019} in multilingual speech processing.

\section{Acknowledgements}
We thank Xinjian Li, Awni Hannun, Alex Shypula, and Xinyi Zhang for helpful discussions. This work was supported in part by grants from National Science Foundation for Bridges PSC (ACI-1548562, ACI-1445606) and DARPA KAIROS program from the Air Force Research Laboratory (FA8750-19-2-0200). The U.S. Government is authorized to reproduce and distribute reprints for Governmental purposes notwithstanding any copyright notation thereon.

\bibliographystyle{IEEEtran}
\bibliography{allograph}

\appendix
\section{Appendix}
\label{sec:appendix}

\subsection{Qualitative Examples of Universal Phone Recognition}
\label{sec:qualitative}

In \Tref{tab:qualitative}, we show qualitative examples of phone transcriptions on two unseen languages along with the phone error rate (PER), substitution error rate (SER), and articulatory feature distance (AFD). As discussed in \Sref{sec:afd}, the AlloGraph models produce intelligible results while the AlloMatrix baseline frequently substitutes vowels for plosives, resulting in high AFD and phone transcriptions that are mostly uninterpretable.  

\begin{table}[h]
  \centering
    \caption{Qualitative examples of universal phone transcriptions of the AlloMatrix baseline and AlloGraph models on two unseen languages, Tusom and Inuktitut. The errors of each phone output sequence are highlighted in \textcolor{red}{red}. The phone error rate (PER), substitution error rate (SER), and articulatory feature distance (AFD) of each sequence are also shown.}
    \label{tab:qualitative}
  \vspace{-5pt}
    \resizebox {\linewidth} {!} {
\begin{tabular}{l|lccc}
\toprule
\multicolumn{5}{c}{\underline{\textsc{Unseen Language}}: Tusom} \\
\midrule
Model / Source & Phone Output & PER & SER & AFD \\
\midrule
AlloMatrix & \textipa{[\textcolor{red}{s's'\textlowering{\textbeta}}]} & 100.0 & 60.0 & 13.3 \\
AlloGraph & \textipa{[\textcolor{red}{\textschwa k}\textbari \textcolor{red}{\textturnr}u]} & 80.0 & 60.0 & 4.7 \\
+ UC & \textipa{[\textglotstop \textbari \textcolor{red}{k}ru]} & 20.0 & 20.0 & 2.0 \\
Ground-Truth & \textipa{[\textglotstop \textbari k\textsuperscript{h}ru]} & - & - & - \\
\midrule

AlloMatrix & \textipa{[b\textcolor{red}{s'\textlowering{\textbeta}\textsubumlaut{g}s'\textturnr}]} & 83.3 & 83.3 & 12.2 \\
AlloGraph & \textipa{[b\textcolor{red}{\textturna}N\textcolor{red}{\textsubumlaut{g}s'\textturnr}]} & 66.6 & 66.6 & 8.3  \\
+ UC & \textipa{[b\textcolor{red}{\textturna}N\textcolor{red}{\textsubumlaut{g}Y}r]} & 50.0 & 50.0 & 4.0 \\
Ground-Truth & \textipa{[baNgor]} & - & - & - \\
\midrule

AlloMatrix & \textipa{[\textcolor{red}{\textlowering{\textbeta}}k\textcolor{red}{s'bs'\textlowering{\textbeta}}]} & 90.0 & 50.0 & 15.4 \\
AlloGraph & \textipa{[\textglotstop \textcolor{red}{o}k\textcolor{red}{u:bu:}S\textcolor{red}{e:}]} & 70.0 & 50.0 & 5.6 \\
+ UC & \textipa{[\textglotstop \textcolor{red}{o}ku\textcolor{red}{bu:}S\textcolor{red}{e:}]} & 60.0 & 40.0 & 6.5 \\
Ground-Truth & \textipa{[\textglotstop ukxuk\textschwa Sue]} & - & - & - \\

\midrule
\midrule

\multicolumn{5}{c}{\underline{\textsc{Unseen Language}}: Inuktitut} \\
\midrule
Model / Source & Phone Output & PER & SER & AFD \\
\midrule
AlloMatrix & \textipa{[k\textcolor{red}{s'Bs'}k k\textcolor{red}{s'Bs'}k]} & 60.0 & 60.0 & 18.3 \\
AlloGraph & \textipa{[ki\textcolor{red}{m}u\textcolor{red}{ck\textsuperscript{h}} ki\textcolor{red}{m}u]} & 50.0 & 30.0 & 6.0 \\
+ UC & \textipa{[k\textcolor{red}{I}N\textcolor{red}{o}k k\textcolor{red}{I}Nuk]} & 30.0 & 30.0 & 2.7 \\
Ground-Truth & \textipa{[kiNuk kiNuk]} & - & - & - \\
\midrule

AlloMatrix & \textipa{[\textcolor{red}{SBs'}k \textcolor{red}{SBks'}]} & 80.0 & 70.0 & 9.7 \\
AlloGraph & \textipa{[s\textcolor{red}{\textbari}k\textcolor{red}{a:k} s\textcolor{red}{u:}k\textcolor{red}{a:k}]} & 60.0 & 60.0 & 2.3 \\
+ UC & \textipa{[\textcolor{red}{\textsubbridge{s}}uk\textcolor{red}{\textturnv k} suk\textcolor{red}{\textturnv k}]} & 50.0 & 50.0 & 2.8 \\
Ground-Truth & \textipa{[sukaq sukaq]} & - & - & - \\
\midrule

AlloMatrix & \textipa{[\textcolor{red}{s'}k\textcolor{red}{s't\textglotstop} \textcolor{red}{s'}k\textcolor{red}{s't}]} & 87.5 & 75.0 & 13.8 \\
AlloGraph & \textipa{[\textcolor{red}{i:}k\textcolor{red}{i:k\textsuperscript{h}} \textcolor{red}{i:}k\textcolor{red}{i:k\textsuperscript{h}}]} & 75.0 & 75.0 & 2.7 \\
+ UC & \textipa{[ik\textcolor{red}{Ip} ik\textcolor{red}{Ipq}]} & 62.5 & 50.0 & 6.5 \\
Ground-Truth & \textipa{[ikiq ikiq]} & - & - & - \\
\bottomrule
\end{tabular}
}
  \vspace{-10pt}
\end{table}

\end{document}